\pdfoutput=1
\documentclass{article}

\usepackage[preprint]{neurips_2026}

\usepackage[utf8]{inputenc} % allow utf-8 input
\usepackage[T1]{fontenc}    % use 8-bit T1 fonts
\usepackage{hyperref}       % hyperlinks
\usepackage{url}            % simple URL typesetting
\usepackage{booktabs}       % professional-quality tables
\usepackage{amsfonts}       % blackboard math symbols
\usepackage{nicefrac}       % compact symbols for 1/2, etc.
\usepackage{microtype}      % microtypography
\usepackage{xcolor}         % colors

\usepackage{graphicx}
\usepackage{amsmath}
\usepackage{amssymb}
\usepackage{enumitem}
\usepackage{placeins}
\usepackage{float}

\title{HALO: Hybrid Adaptive Latent Reasoning for Language Models}

\author{%
Micah Zhang \\
Lockheed Martin \\
\texttt{micah.t.zhang@lmco.com} \\
}

\begin{document}

\maketitle

%%%%%%%%%%%%%%%%%%%%%%%%%%%%%%%%%%%%%%%%%%%%%%%%%%%%%%%%%%%%

\begin{abstract}
We study how to improve a frozen pretrained language model with a small amount of adaptive extra computation. A simple approach is to add additional refinement steps on top of the backbone hidden states, but fixed extra refinement can be wasteful: a one-step refinement head may be too weak, while forcing a second full-sequence refinement step everywhere can increase compute without improving transfer. We introduce HALO, a hybrid adaptive latent-refinement method that combines a coarse refinement stage with selective second-stage latent refinement on a subset of tokens chosen by token scoring and monotonic token halting. On the main public benchmark comparison built from MMLU-Pro and GPQA-Diamond, HALO achieves the best overall average among the paper-facing methods, outperforming the frozen backbone, fixed-1, and fixed-2. Internal analysis further shows that HALO reaches nearly the same token-accuracy level as fixed-2 while using fewer average applied refine steps than fixed-1 and far fewer than fixed-2. These results suggest that the key advantage is not simply more refinement, but a better allocation of refinement: HALO achieves the strongest paper-facing result while also using less measured controller compute than either fixed baseline.
\end{abstract}

%%%%%%%%%%%%%%%%%%%%%%%%%%%%%%%%%%%%%%%%%%%%%%%%%%%%%%%%%%%%

\section{Introduction}

Frozen pretrained language models are strong general-purpose systems, but some reasoning problems benefit from additional inference-time computation. A natural way to add such computation is to refine backbone hidden states after the main forward pass. This can improve reasoning behavior without full-model finetuning, but the extra computation must be allocated carefully.

We introduce \textbf{HALO} (\textbf{H}ybrid \textbf{A}daptive \textbf{L}atent reas\textbf{O}ning), a lightweight adaptive latent-refinement method for frozen language models. HALO uses a coarse refinement stage together with token scoring and monotonic token halting to choose a subset of tokens for a second-stage latent refinement block. Skipped tokens bypass this expensive path and are merged back afterward. The resulting design concentrates the more expensive second-stage computation on only part of the sequence rather than spending a uniform refinement budget everywhere. Importantly, the coarse refinement stage is an architectural path, whereas the paper's compute metric counts the refinement updates actually executed by the controller at runtime. In the winning configuration, these are not the same thing: the initial adaptive keep decision is made token by token from cheap gate features computed from the current logits and, when available, the change from the previous step's logits, so the controller can skip refinement before the later budgeted token-halting stage is reached.

We study a focused question: when extending a frozen language model with extra latent computation, is selective second-step refinement a better use of compute than either stopping after one refinement step or forcing a second refinement step everywhere? We compare HALO against three paper-facing baselines: the frozen backbone, a one-step full-sequence refinement baseline (fixed-1), and a two-step full-sequence refinement baseline (fixed-2).

Our results support a clear answer. On the public comparison built from MMLU-Pro and GPQA-Diamond, HALO achieves the best overall average among the paper-facing methods. Internally, it reaches nearly the same token-accuracy level as fixed-2 while using fewer average applied refine steps than fixed-1 and far fewer than fixed-2. In other words, HALO achieves the strongest paper-facing performance while also using less measured controller compute than either fixed baseline. The gain is not uniform across benchmarks---HALO's strongest advantage comes from GPQA-Diamond while MMLU-Pro remains strongest for the frozen backbone---so the paper makes a focused quality--compute claim rather than a claim of universal dominance. HALO does not win every metric or benchmark column: fixed-2 is marginally strongest on the internal token-accuracy metric, and the frozen backbone remains strongest on MMLU-Pro.

Our contributions are:
\begin{itemize}[leftmargin=1.5em]
    \item We introduce HALO, a hybrid adaptive latent-refinement framework for extending frozen language models with selective second-stage computation.
    \item We show that HALO achieves the best overall public average among the paper-facing methods on MMLU-Pro and GPQA-Diamond.
    \item We show that HALO achieves the strongest paper-facing public result while using fewer average applied refine steps than fixed-1 and far fewer than fixed-2.
\end{itemize}

%%%%%%%%%%%%%%%%%%%%%%%%%%%%%%%%%%%%%%%%%%%%%%%%%%%%%%%%%%%%

\section{Method}

\subsection{Problem setup}

We extend a frozen pretrained language model with a lightweight trainable refinement module. Given an input sequence, the frozen backbone produces contextual token representations and next-token logits. The goal is to improve prediction quality with a small amount of extra computation without paying the cost of applying the same expensive refinement uniformly to every token. Our design target is therefore a better \emph{quality--compute tradeoff}, not simply a larger refinement head.

\subsection{HALO}

\begin{figure}[t]
  \centering
  \includegraphics[width=0.92\linewidth]{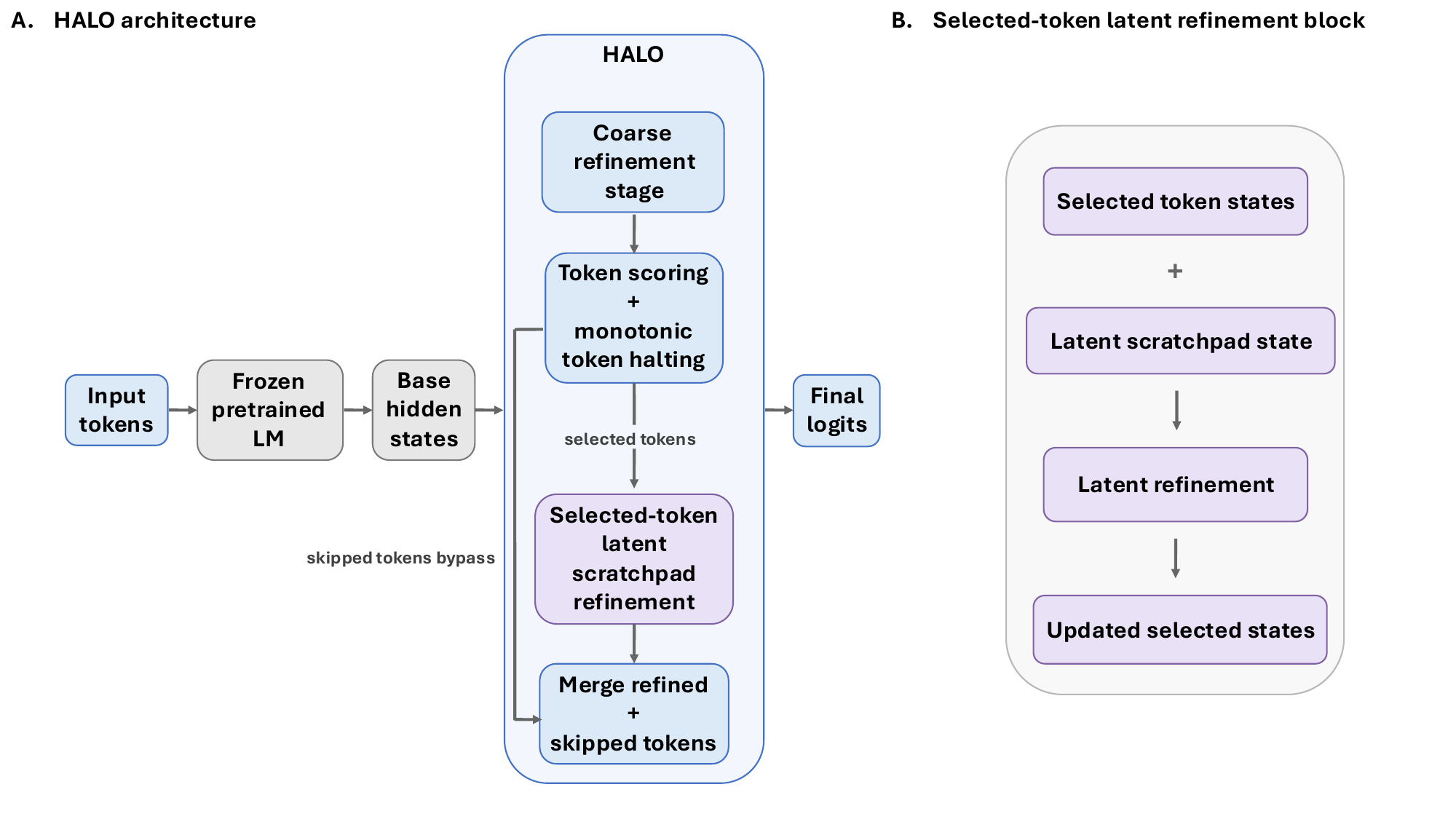}
  \caption{HALO architecture. \textbf{Left:} overall model pipeline. A frozen pretrained language model produces base hidden states, which are passed to HALO before final prediction. HALO uses a coarse refinement stage together with token scoring and monotonic token halting to decide where additional computation is needed. In the winning evaluated configuration, the first adaptive keep decision is made token by token from gate features derived from the current logits and the change from the previous step's logits, so the coarse refinement stage denotes an architectural path rather than a universally forced first update on every token. Selected tokens undergo latent scratchpad refinement, while bypassed tokens skip the expensive path and are merged back before producing final logits. \textbf{Right:} selected-token latent refinement block. Selected token states are combined with a latent scratchpad state and updated through a latent refinement step to produce refined selected representations. This design concentrates extra computation on a subset of tokens rather than forcing additional full-sequence refinement everywhere.}
  \label{fig:halo_architecture}
\end{figure}

Figure~\ref{fig:halo_architecture} shows the overall architecture. Starting from the frozen backbone hidden states, HALO enters a coarse refinement stage. In the winning evaluated configuration, this is an architectural refinement path rather than a universally applied first update on every token. At each refinement step, the model computes token-level gate features from the current logits and, when available, the change from the previous step's logits; a small gate head converts those features into a keep probability, and a token is refined only if that keep probability exceeds threshold and its prediction is not already stable. Because the minimum forced refinement count is zero, the controller can skip the first adaptive refinement update on some token positions.

HALO then performs token scoring and monotonic token halting. In the winning configuration, this selected-token routing is separate from the earlier first-step keep decision. After the coarse stage, HALO assigns a score to candidate tokens for the selected-token path and applies a budgeted monotonic halting rule: only the top-scoring fraction of eligible tokens, subject to a minimum token count, are routed into the expensive second stage. The halting rule is monotonic: once a token is dropped from further refinement within a forward pass, it is not reactivated later. Tokens that are not selected therefore bypass the expensive second stage entirely.

The winning configuration uses a learned-gain token score together with a budget fraction of $0.35$ and a minimum of $8$ selected tokens. Operationally, the controller forms the eligible-token set, scores those tokens, and routes only the top-scoring budgeted fraction into the second stage while still enforcing the minimum token count. Intuitively, this score is meant to prioritize tokens for which an additional latent update is predicted to be most useful, rather than spending the same second-stage budget everywhere. This implements the selective-compute intuition behind HALO by concentrating the second-stage budget on a subset of token positions.

A second important design choice is train--test consistency. Here this means that the same selected-token routing rule used in adaptive evaluation is also applied when fixed evaluation invokes the second-stage refinement path, rather than sending all eligible tokens through a different second-stage regime. In the winning configuration, the same budgeted learned-gain token-halting rule is applied in fixed evaluation as well. In practice, this consistency matters because it makes the selective second-step computation behave like a genuine mechanism rather than a training-only regularizer.

\subsection{Selected-token latent refinement}

For the subset of tokens chosen by the halting mechanism, HALO maintains two representations: a selected-token state and a latent scratchpad state. The selected-token block updates these jointly, then writes the refined selected-token state back into the sequence. The latent scratchpad acts as a small auxiliary working memory for the routed tokens, enabling a richer update than a plain one-step residual transformation while keeping the refinement module lightweight relative to the frozen backbone. In the implementation used here, this selected-token refiner is a compact latent refinement block with a single latent recursion rather than a long inner recurrent rollout.

\subsection{Inference modes}

We evaluate HALO against three paper-facing baselines that differ only in how much additional refinement they use.

\paragraph{Frozen backbone.}
The frozen backbone performs no extra refinement.

\paragraph{Fixed-1.}
The one-step single-state baseline applies one full-sequence refinement step to all tokens.

\paragraph{Fixed-2.}
The two-step single-state baseline applies a second full-sequence refinement step everywhere.

\paragraph{HALO (adaptive).}
HALO uses a coarse refinement stage together with token scoring and monotonic token halting, but allocates the latent second-stage refinement only to selected tokens, while bypassed tokens skip that path and are merged back afterward. In the winning adaptive configuration, the first-step keep decision is made token by token using a gate probability computed from cheap logit-based features together with a stability check, so the controller can skip the first adaptive refinement update on some tokens before the later selected-token budget is applied. The later budgeted token-halting mechanism is a second, distinct routing rule that restricts which eligible tokens enter the selected-token latent path. As a result, the runtime compute can fall below one average applied refinement update per token.

This comparison is intentionally narrow. Fixed-1 asks whether one uniform refinement step is enough. Fixed-2 asks whether simply adding more full-sequence refinement improves results. HALO asks whether selective latent refinement can turn a limited extra-compute budget into a better quality--compute tradeoff than either fixed alternative.

%%%%%%%%%%%%%%%%%%%%%%%%%%%%%%%%%%%%%%%%%%%%%%%%%%%%%%%%%%%%

\section{Experimental setup}

\subsection{Model family and checkpoints}

All experiments use a frozen pretrained language model as the backbone, extended with lightweight trainable refinement heads rather than full-model finetuning. In the training and evaluation harness, the default frozen backbone is \texttt{microsoft/Phi-4-mini-instruct} \citep{abdin2025phi4mini}. The winning HALO configuration uses the selected-token latent refinement variant together with budgeted token halting, a token-selection budget of $0.35$, and train--test consistency in fixed evaluation.

We train only the refinement/controller parameters while keeping the backbone frozen. Training uses a supervised next-token objective on instruction-tuning data together with auxiliary losses that shape adaptive-compute behavior, including anytime supervision across refinement steps, controller regularization toward a target refinement budget, and consistency training across budget settings. The winning configuration is trained in this frozen-backbone refinement setup on a 4k-example subset of the HuggingFaceH4 UltraChat-200k instruction-tuning data, a filtered derivative of UltraChat \citep{ding2023ultrachat}, with sequence length 1024, batch size 1 with gradient accumulation of 8, one epoch, and learning rate $10^{-4}$.

For the public results, trainable methods are evaluated across six independently trained checkpoints. The benchmark launch matrix runs the frozen backbone once, then evaluates the trainable checkpoints in fixed and adaptive modes on the benchmark suite using lm-eval \citep{biderman2024lmeval}. For the internal analysis, we evaluate the same family of checkpoints with a separate self-contained internal evaluation matrix that reports quality and compute summaries from the model's native evaluation harness.

\subsection{Public and internal evaluation}

Our public evaluation centers on MMLU-Pro and GPQA-Diamond. These are the two tasks used in the public benchmark matrix for the paper-facing comparison, and we treat them as the main transfer test because they probe complementary forms of challenging reasoning while remaining standard enough for direct comparison across methods. We do not claim that two benchmarks exhaustively characterize the mechanism; rather, they provide a focused but challenging public test for the narrow comparison studied in this paper. For trainable methods, we report mean and standard deviation across six seeds; for the frozen backbone, we report a single deterministic score because the backbone itself is not retrained. Following the main-paper comparison, we summarize public performance using MMLU-Pro, GPQA-Diamond, and their simple average. We use the average only as a compact summary statistic for the joint public picture, while keeping the benchmark-specific columns explicit because the methods do not behave identically across the two tasks.

In addition to the public benchmarks, we run an internal evaluation designed to expose the quality--compute tradeoff more directly than lm-eval alone. The internal harness reports token accuracy, token-accuracy lift over the frozen backbone, and average applied refine steps, along with related adaptive and fixed evaluation summaries. It evaluates each checkpoint through the model's native fixed and adaptive evaluation modes and records controller-side quantities such as per-token used-step counts, gate statistics, and prediction-delta traces, from which the reported internal quality and compute summaries are derived. The internal evaluation matrix runs the trained checkpoints with up to 512 evaluation samples, sequence length 1024, and batch size 1. We focus on three internal quantities: token accuracy, token-accuracy lift over the frozen backbone, and average applied refine steps as a compute proxy. This metric counts the refinement updates actually executed by the controller at token positions during evaluation rather than a fixed architectural stage count. Operationally, the model maintains a per-token counter of executed refinement updates: a token skipped at all steps contributes $0$, a token refined once contributes $1$, and a token refined twice contributes $2$, and the reported quantity averages that count over predicted token positions. Because the winning HALO configuration allows the first adaptive refinement update to be skipped on some tokens, HALO can fall below 1.0 on this metric even though the architecture contains a coarse refinement stage. Throughout the paper, this is the load-bearing efficiency metric: it is the quantity on which HALO is more compute-efficient than both fixed baselines. We use average applied refine steps as the primary compute proxy because it is measured directly by the model's own refinement controller and is available consistently across fixed and adaptive modes.

\subsection{Baselines and reporting}

The paper compares four methods: the frozen backbone, single-state fixed-1, single-state fixed-2, and HALO (adaptive). These baselines are intentionally narrow: the goal is not a broad family sweep, but a focused test of whether selective second-step latent refinement is a better use of compute than either stopping after one full-sequence refinement step or forcing a second step everywhere.

All table values in the paper are taken from the consolidated verified results workbook used to merge the public and internal evaluations into a single checked artifact. The frozen backbone is reported once as a deterministic reference; the trainable public and internal rows report means and standard deviations across the six trainable seeds. Average applied refine steps are reported as the main method-level compute proxy.

\subsection{Reporting protocol}

The public benchmark matrix evaluates the frozen backbone in teacher mode and the trainable checkpoints in fixed and adaptive modes. The internal evaluation matrix analogously re-evaluates the trained checkpoints through the model's own evaluation path. Overall, the experimental setup is designed to make the comparison as controlled as possible: the frozen backbone is shared, the refinement families are evaluated under matched benchmark protocols, and the main comparison isolates whether selective latent refinement provides a better quality--compute tradeoff than fixed one-step or fixed two-step alternatives.

%%%%%%%%%%%%%%%%%%%%%%%%%%%%%%%%%%%%%%%%%%%%%%%%%%%%%%%%%%%%

\section{Results}

We evaluate HALO against three paper-facing comparators: the frozen backbone, a one-step single-state baseline (fixed-1), and a two-step single-state baseline (fixed-2). The public-benchmark results establish the main transfer claim, while the internal results clarify the quality--efficiency tradeoff behind that claim. The central empirical result of the paper is that HALO achieves the strongest paper-facing public average while also using fewer average applied refine steps than fixed-1 and far fewer than fixed-2. Across these experiments, the central question is therefore not whether extra refinement always helps, but whether \emph{selective} extra refinement yields a better quality--compute tradeoff than simply forcing additional full-sequence refinement everywhere.

\subsection{Public benchmark results}

\begin{table}[t]
\caption{Public benchmark results on MMLU-Pro and GPQA-Diamond. We report mean $\pm$ standard deviation across six seeds for trainable methods and a single deterministic score for the frozen backbone. HALO achieves the best overall average, outperforming all paper-facing baselines while avoiding the extra full-sequence second step of fixed-2. The gains are driven primarily by GPQA-Diamond, whereas MMLU-Pro remains strongest for the frozen backbone.}
\label{tab:public_benchmarks}
\centering
\small
\begin{tabular}{lccc}
\toprule
Method & MMLU-Pro $\uparrow$ & GPQA-Diamond $\uparrow$ & Average $\uparrow$ \\
\midrule
Frozen backbone & \textbf{38.54} & 31.82 & 35.18 \\
Single-state baseline (fixed-1) & 38.49 $\pm$ 0.09 & 32.32 $\pm$ 2.02 & 35.41 $\pm$ 1.01 \\
Single-state baseline (fixed-2) & 38.43 $\pm$ 0.12 & 32.07 $\pm$ 1.65 & 35.25 $\pm$ 0.78 \\
HALO (adaptive) & 38.33 $\pm$ 0.37 & \textbf{33.00 $\pm$ 0.99} & \textbf{35.66 $\pm$ 0.43} \\
\bottomrule
\end{tabular}
\end{table}

Table~\ref{tab:public_benchmarks} presents the main public-benchmark results on MMLU-Pro and GPQA-Diamond. HALO achieves the best overall average, improving over the frozen backbone, the one-step single-state baseline, and the two-step single-state baseline.

The pattern across benchmarks is asymmetric. On MMLU-Pro, the frozen backbone remains strongest at $38.54$, while HALO reaches $38.33 \pm 0.37$. The public gain therefore does not come from uniform improvement across all benchmarks, and the paper should not be read as claiming that it does. Instead, it is driven primarily by GPQA-Diamond, where HALO reaches $33.00 \pm 0.99$, outperforming the frozen backbone ($31.82$), fixed-1 ($32.32 \pm 2.02$), and fixed-2 ($32.07 \pm 1.65$).

The comparison between fixed-1 and fixed-2 is also informative. If the benefit of HALO were mainly due to using more refinement steps, then forcing a second full-sequence refinement step everywhere should have improved the single-state baseline. Instead, fixed-2 performs slightly worse than fixed-1 on the overall public average ($35.25 \pm 0.78$ versus $35.41 \pm 1.01$). This suggests that the key advantage is not simply ``more refinement,'' but rather \emph{where} and \emph{how} the additional computation is applied. HALO allocates second-step computation selectively, whereas fixed-2 pays for an unconditional second full-sequence pass.

Overall, the public results support a clear conclusion: HALO provides the best paper-facing transfer result on this focused public benchmark pair while leaving the benchmark-level asymmetry visible rather than hiding it.

\subsection{Internal quality--efficiency comparison}

\begin{table}[t]
\caption{Internal quality and efficiency comparison for the paper-facing methods. We report internal token accuracy, token-accuracy lift over the frozen backbone, and average applied refine steps across six seeds for the trainable methods. Average applied refine steps count the refinement updates actually executed by the controller. Among trainable methods, HALO uses fewer applied refinement updates than fixed-1 and much less compute than fixed-2, while fixed-2 is only marginally higher than HALO on this internal token-accuracy metric.}
\label{tab:internal_quality_efficiency}
\centering
\small
\begin{tabular}{lccc}
\toprule
Method & Token acc. $\uparrow$ & Lift vs.\ frozen $\uparrow$ & Avg.\ applied refine steps $\downarrow$ \\
\midrule
Frozen backbone & 0.6960 $\pm$ 0.0000 & 0.0000 $\pm$ 0.0000 & 0.000 $\pm$ 0.000 \\
Single-state baseline (fixed-1) & 0.7050 $\pm$ 0.0012 & 0.0089 $\pm$ 0.0012 & 1.000 $\pm$ 0.000 \\
Single-state baseline (fixed-2) & \textbf{0.7067 $\pm$ 0.0008} & \textbf{0.0107 $\pm$ 0.0008} & 2.000 $\pm$ 0.000 \\
HALO (adaptive) & 0.7066 $\pm$ 0.0004 & 0.0106 $\pm$ 0.0004 & \textbf{0.776 $\pm$ 0.009} \\
\bottomrule
\end{tabular}
\end{table}

Table~\ref{tab:internal_quality_efficiency} complements the public results by comparing internal quality and compute for the same paper-facing methods. Here the picture is more nuanced, and that nuance is useful. The fixed-2 baseline achieves the highest internal token accuracy at $0.7067 \pm 0.0008$, with HALO essentially tied at $0.7066 \pm 0.0004$. The absolute difference between them is extremely small. By contrast, the compute difference is large: fixed-2 uses exactly $2.000$ average applied refine steps, whereas HALO uses only $0.776 \pm 0.009$.

This comparison is central to the paper's interpretation. On the internal metric, fixed-2 can match or slightly exceed HALO in raw token accuracy, but it does so by always paying for a second full-sequence refinement step. HALO reaches nearly the same internal quality while using far less additional computation. More strikingly, HALO also uses fewer average applied refine steps than fixed-1. This is possible because the reported compute metric counts executed token-level refinement updates, not architectural stages. In the winning HALO configuration, the first adaptive refinement update is not forced on every token: the controller computes a keep probability from cheap logit-based features and skips tokens whose keep probability stays below threshold or whose predictions are already stable. In other words, HALO is not best understood as maximizing the internal metric at all costs; rather, it offers a substantially more efficient way to obtain nearly the same internal quality.

The internal results also help contextualize fixed-1. Relative to the frozen backbone, fixed-1 improves token accuracy by $0.0089 \pm 0.0012$ with exactly one average applied refine step, while fixed-2 improves by $0.0107 \pm 0.0008$ with two average applied refine steps. HALO improves by $0.0106 \pm 0.0004$ while using less than one average applied refine step. Thus, HALO clearly exceeds the one-step baseline in internal quality while also using less measured controller compute, while remaining far below the two-step baseline in compute.

Taken together, Table~\ref{tab:internal_quality_efficiency} supports the paper's core efficiency claim. A simple second full-sequence step can be competitive on the internal metric, but HALO reaches comparable internal quality with much lower average compute and translates that tradeoff into the strongest public average.

\subsection{Public quality versus compute}

\begin{figure}[t]
  \centering
  \includegraphics[width=0.74\linewidth]{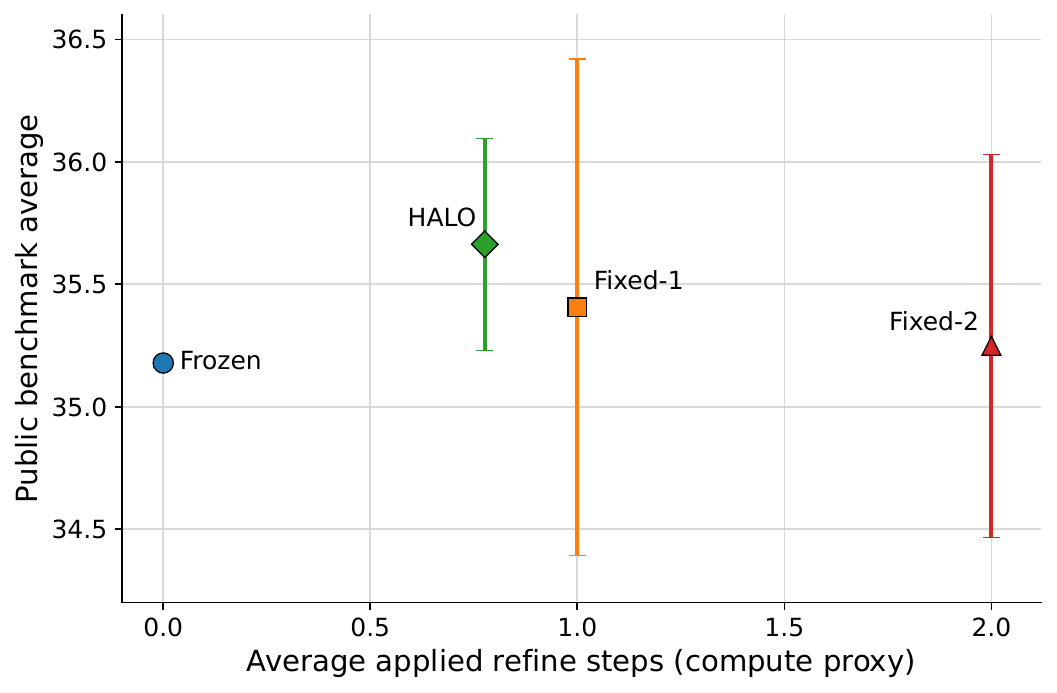}
  \caption{Public quality versus compute. The y-axis shows the average of MMLU-Pro and GPQA-Diamond from the verified public benchmark table, and the x-axis shows average applied refine steps as a compute proxy, i.e. the average number of refinement updates actually executed by the controller. Error bars show variation across seeds. HALO achieves higher public-benchmark average than the one-step single-state baseline while using fewer applied refinement updates than fixed-1 and far fewer than fixed-2, and it exhibits lower seed-to-seed variability than the single-state baselines.}
  \label{fig:public_quality_vs_compute}
\end{figure}

Figure~\ref{fig:public_quality_vs_compute} summarizes the public quality--compute tradeoff visually. The x-axis shows average applied refine steps as a compute proxy, and the y-axis shows the average of MMLU-Pro and GPQA-Diamond from Table~\ref{tab:public_benchmarks}. The frozen backbone anchors the zero-compute point. Fixed-1 is weaker than HALO and, under this controller-compute metric, also uses more applied refinement than HALO. Fixed-2 uses substantially more compute but does not improve the public average over fixed-1. HALO occupies the most attractive region of the plot: it attains the best public average while lying to the left of both fixed baselines.

This figure makes the paper's main systems-style point especially clear. If the goal were only to maximize public accuracy with no regard for compute, then one might expect the best method to sit at the far right of the curve, where the most computation is spent. Instead, the two-step single-state baseline shows that simply spending more full-sequence compute is not enough. HALO's advantage comes from \emph{selective} computation: it improves over fixed-1 while using fewer applied refinement updates than fixed-1, and without inheriting the cost profile of fixed-2. Because the public average is computed from only two benchmarks, the figure should be read as a focused quality--compute summary rather than a universal frontier over all tasks.

Figure~\ref{fig:public_quality_vs_compute} also reinforces the variance pattern already visible in Table~\ref{tab:public_benchmarks}. The HALO point is accompanied by smaller error bars than the single-state baselines, indicating lower seed-to-seed variation in the public average.

\subsection{Summary of the main finding}

Across the public and internal evaluations, a consistent picture emerges. The one-step single-state baseline is weaker than HALO; the two-step single-state baseline spends more compute but does not improve the public average; and HALO achieves the strongest public result while using fewer average applied refine steps than fixed-1 and substantially less compute than fixed-2. This is the paper's central result: HALO is simultaneously the best-performing paper-facing method on the main public average and the most compute-efficient of the trainable paper-facing methods under the reported controller-compute metric. The most important takeaway is therefore not that more refinement is always better, but that \emph{selective latent refinement} is a more effective use of extra computation than applying additional full-sequence refinement everywhere.

\subsection{Discussion and limitations}

The results support a specific claim about \emph{how} extra computation should be used in this setting. HALO is not strongest because it wins every metric or every benchmark column. Rather, it delivers the best overall public transfer result while using substantially less compute than the unconditional two-step baseline. The public improvement is asymmetric: HALO's gain is driven primarily by GPQA-Diamond, while MMLU-Pro remains strongest for the frozen backbone.

The internal results sharpen the same point. Fixed-2 is marginally stronger than HALO on raw internal token accuracy, but only by a very small amount and at much higher compute. HALO therefore should not be read as maximizing the internal metric at all costs; its value lies in reaching nearly the same internal quality with far fewer average applied refine steps and converting that efficiency advantage into the strongest public average.

This also clarifies one of the most important negative results in the paper. If HALO's gain came primarily from simply adding another refinement step, then fixed-2 should have been a stronger public baseline than fixed-1. Instead, fixed-2 spends more computation than fixed-1 by applying a second full-sequence refinement step everywhere, yet it does not improve the public average. The public results therefore suggest that indiscriminately spending more full-sequence compute is not enough. What matters is whether the additional computation is targeted.

The internal and public results should not be treated as interchangeable. The internal metric exposes quality and compute with fine granularity, but the public benchmarks are load-bearing because they test whether the refinement mechanism transfers beyond the model's native harness.

Our evidence also has clear limits. The public story is built on two benchmarks, the overall win is modest, and the main comparison is intentionally narrow. The benchmark-level gain is also asymmetric, with the strongest advantage coming from GPQA-Diamond rather than from a uniform improvement across both public tasks. We therefore present HALO as evidence that selective latent refinement can improve the quality--compute tradeoff of frozen language models, not as evidence that adaptive reasoning for frozen language models is solved.

A broader implication of HALO is that better quality--compute efficiency could reduce the deployment cost of language-model reasoning for beneficial applications. At the same time, improved inference efficiency may also make capable generative systems cheaper to run and easier to scale, which could amplify misuse risks already associated with large language models. We therefore view HALO as a methods contribution whose societal effects depend on the downstream deployment context of the underlying pretrained model.

These limitations point naturally to future directions. One next step is to test whether the same selective refinement mechanism holds up across a broader benchmark set, including tasks with different mixtures of recall, structured reasoning, and long-context dependence. Another is to measure efficiency more directly, for example through wall-clock latency or throughput in addition to average applied refine steps. A third is to improve the refinement policy itself, since the present results already suggest that where computation is allocated matters more than simply increasing it.

Overall, we view HALO as evidence for a particular design principle. When extending a frozen pretrained language model with extra reasoning compute, the key question is not just whether additional refinement helps, but whether that refinement is allocated selectively enough to justify its cost. Our results suggest that the answer is yes: selective latent refinement can outperform a simpler one-step baseline while also using fewer applied refinement updates, and while avoiding the public inefficiency of a fully applied second-step baseline. That is a narrower claim than universal superiority, but it is also the one most directly supported by the evidence in this paper.

%%%%%%%%%%%%%%%%%%%%%%%%%%%%%%%%%%%%%%%%%%%%%%%%%%%%%%%%%%%%

\section{Related work}

HALO is related to several lines of work on adaptive computation, recurrent refinement, token-level halting, and frozen-model extension.

\paragraph{Adaptive computation.}
Adaptive Computation Time, Universal Transformers, and PonderNet study how neural sequence models can allocate variable amounts of computation rather than using a fixed budget everywhere \citep{graves2016adaptive,dehghani2019universal,banino2021pondernet}. HALO belongs to this broad tradition, but targets frozen language-model extension rather than end-to-end recurrent sequence modeling.

\paragraph{Recurrent and latent refinement.}
Recent recurrent-depth and latent-reasoning language models argue that repeated latent updates can provide a form of inference-time scaling distinct from explicit chain-of-thought generation \citep{geiping2025scaling,hao2024continuous,saunshi2025looped,zeng2025ponderlm}. Closely related small-model work such as TRM also studies recursive latent refinement through repeated latent-state and answer-state updates \citep{jolicoeurmartineau2025trm}. HALO is related in spirit to this line of work, but differs in focusing on a lightweight refinement head on top of a frozen pretrained backbone rather than recurrently training the full model.

\paragraph{Selective computation and token-level halting.}
Mixture-of-Depths, early-exit methods, AdaPonderLM, and ANIRA study token-level adaptive depth or halting in Transformers and language models \citep{raposo2024mixture,elhoushi2024layerskip,shan2024earlyexit,song2026adaponderlm,moosa2026understanding}. HALO is most directly aligned with this line of work: its core mechanism is budgeted monotonic token halting that routes only a selected subset of tokens into a second-stage latent refinement block.

\paragraph{Frozen-model refinement.}
HALO is also connected to methods that improve a pretrained language model by adding an auxiliary refinement or self-improvement stage at inference time, including output-space self-refinement and latent-space reasoning \citep{madaan2023selfrefine,hao2024continuous,geiping2025scaling,zeng2025ponderlm}. HALO differs in its emphasis on \emph{selective} hidden-state refinement over a frozen backbone.

Overall, HALO sits at the intersection of these threads: learned halting, latent hidden-state refinement, and frozen-model extension with a small trainable module. Our empirical focus is narrower than most related work: a controlled comparison showing that selective second-step latent refinement is a better use of extra compute than unconditional additional full-sequence refinement in the frozen-backbone setting studied here.

%%%%%%%%%%%%%%%%%%%%%%%%%%%%%%%%%%%%%%%%%%%%%%%%%%%%%%%%%%%%

\section{Conclusion}

We introduced HALO, a hybrid adaptive latent-refinement method for extending a frozen pretrained language model with selective extra computation. HALO combines a coarse refinement stage with a selected-token latent refinement stage, allowing the model to spend additional computation only where it appears most useful rather than forcing extra full-sequence refinement everywhere.

Across the public benchmarks used in our primary comparison, HALO achieves the best overall average among the paper-facing methods, outperforming the frozen backbone, fixed-1, and fixed-2 on this focused two-benchmark comparison.

The internal analysis clarifies this tradeoff further. HALO reaches nearly the same internal token-accuracy level as the two-step baseline while using fewer average applied refine steps than the one-step baseline and far fewer than the two-step baseline. Taken together, these results suggest that the key advantage is not merely more refinement, but better allocation of refinement. More broadly, HALO provides focused evidence that selective latent refinement can improve the quality--compute tradeoff of frozen language models.

%%%%%%%%%%%%%%%%%%%%%%%%%%%%%%%%%%%%%%%%%%%%%%%%%%%%%%%%%%%%

\section*{Acknowledgments}
I thank Nathan Rigoni for his support of this project and for giving me the space to pursue this research and develop it into a paper.

%%%%%%%%%%%%%%%%%%%%%%%%%%%%%%%%%%%%%%%%%%%%%%%%%%%%%%%%%%%%

% \section*{References}
{\small
\nocite{biderman2024lmeval,abdin2025phi4mini,phi2024technical,wang2024mmlupro,rein2023gpqa,hendrycks2021mmlu,wei2022chain,snell2024ttc,goyal2023pause,bae2025mor,jeddi2026loopformer,rauba2026tinyautoregressive,sheshanarayana2026adaanchor,zeng2026adaptivelatentcot,chen2025itt}
\bibliographystyle{plainnat}
\bibliography{halo_references}
}

%%%%%%%%%%%%%%%%%%%%%%%%%%%%%%%%%%%%%%%%%%%%%%%%%%%%%%%%%%%%

\clearpage
\appendix

\section{Additional experimental details}

\subsection{Winning HALO configuration}

The appendix is intended to support auditability of the paper's narrow comparison rather than to introduce a broader family sweep. Accordingly, we report only the exact paper-facing adaptive configuration and the corresponding fixed baselines.

The winning HALO configuration used in the paper (internally labeled v139) uses selected-token latent refinement together with budgeted token halting, a token-selection budget of $0.35$, a minimum of $8$ selected tokens, and train--test consistency in fixed evaluation. It also uses multibudget consistency training and a learned token-scoring rule for selecting which tokens receive additional latent refinement. Operationally, the selected-token controller scores eligible tokens, keeps only the top-scoring budgeted fraction subject to the minimum token count, and routes only those tokens into the selected-token latent block. In this configuration, monotonic token halting means that once a token is dropped from further selected-token refinement within a forward pass, it is not reactivated later. Train--test consistency means that this same budgeted learned-gain routing rule is also applied when fixed evaluation invokes the second-stage path.

The paper-facing fixed baselines are intentionally simple. The frozen backbone performs no refinement. The fixed-1 baseline applies one full-sequence single-state refinement step. The fixed-2 baseline applies two full-sequence single-state refinement steps. These rows are included because they answer the paper's central question directly: whether selective second-stage latent refinement is a better use of limited extra computation than either stopping after one refinement step or forcing a second full-sequence refinement step everywhere.

\subsection{Training recipe}

The winning HALO checkpoints are trained in the harness's frozen-backbone refinement setting, which freezes the pretrained backbone and optimizes only the refinement/controller module. The training data path is an instruction-tuning chat setup using the HuggingFaceH4 UltraChat-200k data, a filtered derivative of UltraChat \citep{ding2023ultrachat}, and the winning run uses 4{,}000 training examples, 256 held-out evaluation examples during training, sequence length 1024, batch size 1 with gradient accumulation of 8, one epoch, BF16, activation checkpointing, and learning rate $10^{-4}$.

The training objective is multi-term rather than plain next-token cross-entropy alone. In addition to supervised next-token learning, the winning setup includes anytime supervision across refinement steps, regularization toward a target adaptive-compute budget, gate warm-start/oracle-style controller supervision, and multibudget consistency training. These ingredients matter because HALO's controller behavior is learned rather than hand-specified: they encourage the refinement head to improve predictions while also learning when to spend or withhold extra computation.

\subsection{Public benchmark protocol}

The public benchmark comparison is built from MMLU-Pro and GPQA-Diamond. In the public lm-eval matrix, the frozen backbone is evaluated once in teacher mode, and the trainable checkpoints are evaluated in fixed and adaptive modes. The public matrix explicitly lists MMLU-Pro and GPQA-Diamond as the paper-facing tasks, and the main paper reports the frozen backbone, fixed-1, fixed-2, and HALO rows derived from the consolidated verified workbook.

For trainable methods, public results are aggregated across six independently trained checkpoints with seeds 1234, 2234, 3234, 4234, 5234, and 6234. The frozen backbone is deterministic and is therefore reported once. In the paper, we summarize public performance with MMLU-Pro, GPQA-Diamond, and their simple average. Table~\ref{tab:appendix_public_per_seed} reports the underlying per-seed values used to form Table~\ref{tab:public_benchmarks}.

\subsection{Internal evaluation protocol}

The internal evaluation uses the model's native evaluation path. The self-contained internal evaluation matrix re-evaluates the paper-facing fixed-baseline and HALO checkpoints across the same six seeds used in the public table. Each call uses up to 512 evaluation samples, sequence length 1024, and batch size 1. As in the main text, we focus on token accuracy, token-accuracy lift over the frozen backbone, and average applied refine steps as the primary method-level compute proxy.

Average applied refine steps count the refinement updates actually executed by the controller rather than a fixed architectural stage count. This is the quantity used in the paper's quality--compute analysis and in the appendix quality--compute scatter.

\subsection{Compute resources and external assets}

Training and evaluation runs used a single NVIDIA A100 GPU per run; multiple runs were sometimes executed in parallel, but each individual model was trained and evaluated on one A100. The experiments use the MIT-licensed Phi-4-mini-instruct model, the MIT-licensed lm-evaluation-harness, the Apache-2.0-licensed MMLU-Pro repository, GPQA assets released under CC BY 4.0 for the dataset card and MIT for the code repository, and the HuggingFaceH4 UltraChat-200k training data released under the OpenRAIL license.

\section{Per-seed results}

\subsection{Per-seed public benchmark results}

Table~\ref{tab:appendix_public_per_seed} reports the per-seed public benchmark values underlying the aggregated results in the main public table.

\begin{table}[H]
\caption{Per-seed public benchmark results for the paper-facing methods. Values are percentages. The frozen backbone is deterministic and is reported once. This table is the appendix companion to Table~\ref{tab:public_benchmarks}: it makes the reported means and standard deviations directly auditable.}
\label{tab:appendix_public_per_seed}
\centering
\small
\begin{tabular}{llccc}
\toprule
Method & Seed & MMLU-Pro $\uparrow$ & GPQA-Diamond $\uparrow$ & Average $\uparrow$ \\
\midrule
Frozen backbone & — & 38.54 & 31.82 & 35.18 \\
Single-state baseline (fixed-1) & 1234 & 38.38 & 32.83 & 35.60 \\
Single-state baseline (fixed-1) & 2234 & 38.47 & 28.79 & 33.63 \\
Single-state baseline (fixed-1) & 3234 & 38.41 & 33.33 & 35.87 \\
Single-state baseline (fixed-1) & 4234 & 38.50 & 31.82 & 35.16 \\
Single-state baseline (fixed-1) & 5234 & 38.62 & 32.32 & 35.47 \\
Single-state baseline (fixed-1) & 6234 & 38.55 & 34.85 & 36.70 \\
Single-state baseline (fixed-2) & 1234 & 38.35 & 33.84 & 36.09 \\
Single-state baseline (fixed-2) & 2234 & 38.47 & 31.82 & 35.15 \\
Single-state baseline (fixed-2) & 3234 & 38.42 & 33.33 & 35.88 \\
Single-state baseline (fixed-2) & 4234 & 38.23 & 32.83 & 35.53 \\
Single-state baseline (fixed-2) & 5234 & 38.53 & 31.31 & 34.92 \\
Single-state baseline (fixed-2) & 6234 & 38.56 & 29.29 & 33.92 \\
HALO (adaptive) & 1234 & 38.21 & 32.32 & 35.27 \\
HALO (adaptive) & 2234 & 37.82 & 33.33 & 35.58 \\
HALO (adaptive) & 3234 & 38.05 & 34.34 & 36.20 \\
HALO (adaptive) & 4234 & 38.83 & 31.82 & 35.32 \\
HALO (adaptive) & 5234 & 38.59 & 33.84 & 36.21 \\
HALO (adaptive) & 6234 & 38.47 & 32.32 & 35.40 \\
\bottomrule
\end{tabular}

\end{table}

\begin{figure}[H]
  \centering
  \includegraphics[width=0.74\linewidth]{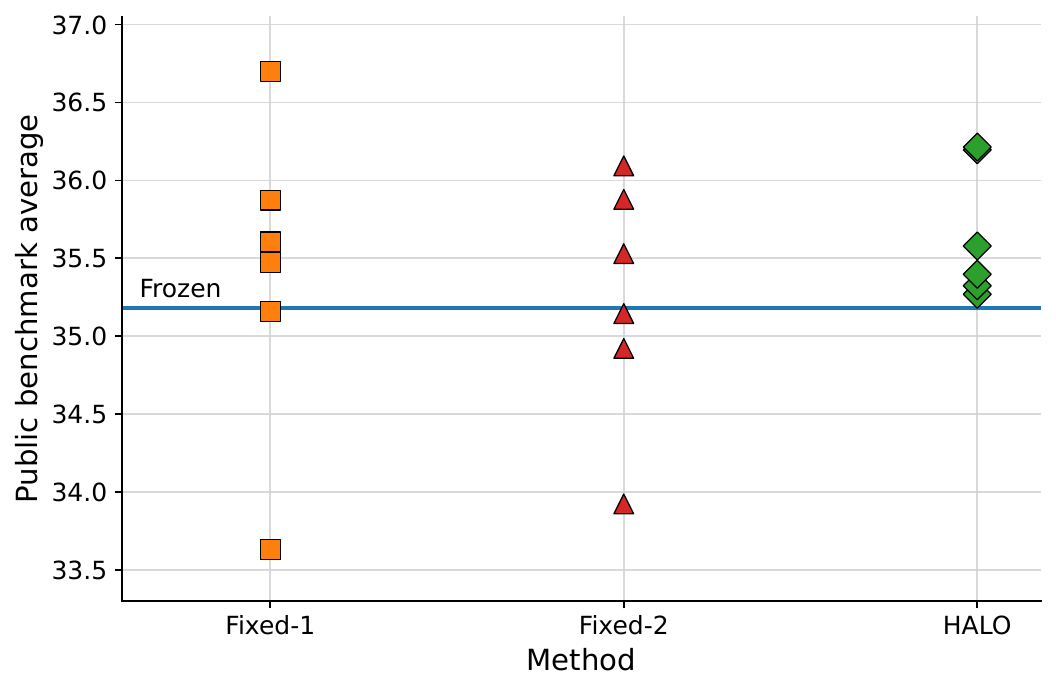}
  \caption{Seed-level public benchmark averages for the paper-facing methods. Each point corresponds to one seed for a trainable method, and the frozen backbone is shown as a reference. This plot is the visual companion to Table~\ref{tab:appendix_public_per_seed} and makes the public variance pattern easier to inspect.}
  \label{fig:appendix_public_seed_average}
\end{figure}

Figure~\ref{fig:appendix_public_seed_average} provides a compact visual view of the per-seed public averages reported in Table~\ref{tab:appendix_public_per_seed}. Across these six seeds, every HALO checkpoint remains above the frozen-backbone public average, whereas both fixed baselines include seeds that fall below the frozen-backbone reference. This is consistent with the smaller public-average standard deviation reported for HALO in Table~\ref{tab:public_benchmarks} and suggests a more stable public transfer gain for HALO on this benchmark pair, but not that the effect would necessarily continue to grow with more training or a larger backbone.

\FloatBarrier

\subsection{Per-seed internal quality and efficiency results}

Table~\ref{tab:appendix_internal_per_seed} reports the per-seed internal quality and efficiency values underlying the aggregated results in Table~\ref{tab:internal_quality_efficiency}.

\begin{table}[H]
\caption{Per-seed internal quality and efficiency results for the paper-facing methods. Average applied refine steps count the refinement updates actually executed by the controller. The frozen backbone is deterministic and is reported once. This table is the appendix companion to Table~\ref{tab:internal_quality_efficiency}.}
\label{tab:appendix_internal_per_seed}
\centering
\small
\begin{tabular}{llccc}
\toprule
Method & Seed & Token acc. $\uparrow$ & Lift vs.\ frozen $\uparrow$ & Avg.\ applied refine steps $\downarrow$ \\
\midrule
Frozen backbone & — & 0.6960 & 0.0000 & 0.000 \\
Single-state baseline (fixed-1) & 1234 & 0.7027 & 0.0067 & 1.000 \\
Single-state baseline (fixed-1) & 2234 & 0.7057 & 0.0097 & 1.000 \\
Single-state baseline (fixed-1) & 3234 & 0.7055 & 0.0095 & 1.000 \\
Single-state baseline (fixed-1) & 4234 & 0.7053 & 0.0093 & 1.000 \\
Single-state baseline (fixed-1) & 5234 & 0.7047 & 0.0087 & 1.000 \\
Single-state baseline (fixed-1) & 6234 & 0.7058 & 0.0098 & 1.000 \\
Single-state baseline (fixed-2) & 1234 & 0.7055 & 0.0095 & 2.000 \\
Single-state baseline (fixed-2) & 2234 & 0.7072 & 0.0112 & 2.000 \\
Single-state baseline (fixed-2) & 3234 & 0.7076 & 0.0116 & 2.000 \\
Single-state baseline (fixed-2) & 4234 & 0.7068 & 0.0108 & 2.000 \\
Single-state baseline (fixed-2) & 5234 & 0.7069 & 0.0109 & 2.000 \\
Single-state baseline (fixed-2) & 6234 & 0.7061 & 0.0101 & 2.000 \\
HALO (adaptive) & 1234 & 0.7061 & 0.0101 & 0.764 \\
HALO (adaptive) & 2234 & 0.7067 & 0.0107 & 0.788 \\
HALO (adaptive) & 3234 & 0.7065 & 0.0104 & 0.784 \\
HALO (adaptive) & 4234 & 0.7065 & 0.0105 & 0.774 \\
HALO (adaptive) & 5234 & 0.7065 & 0.0105 & 0.776 \\
HALO (adaptive) & 6234 & 0.7072 & 0.0112 & 0.773 \\
\bottomrule
\end{tabular}

\end{table}

\begin{figure}[H]
  \centering
  \includegraphics[width=0.66\linewidth]{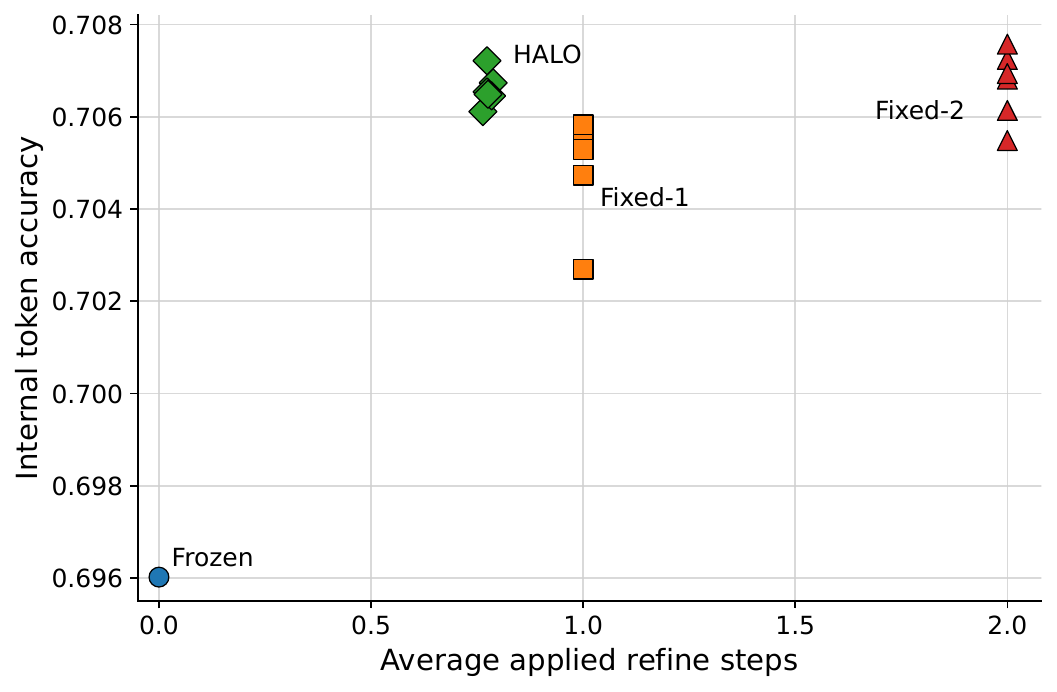}
  \caption{Seed-level internal quality versus compute. Each point corresponds to one seed, the x-axis shows average applied refine steps, and the y-axis shows internal token accuracy. The frozen backbone is shown as a reference. This plot is the seed-level appendix analogue of the paper's compute story: HALO clusters near the fixed-2 token-accuracy region while using fewer applied refinement updates than fixed-1 and much fewer than fixed-2.}
  \label{fig:appendix_internal_quality_vs_compute}
\end{figure}

Figure~\ref{fig:appendix_internal_quality_vs_compute} gives the corresponding seed-level internal quality--compute view.

\FloatBarrier

%%%%%%%%%%%%%%%%%%%%%%%%%%%%%%%%%%%%%%%%%%%%%%%%%%%%%%%%%%%%

\end{document}